# Inference in Probabilistic Logic Programs using Weighted CNF's

**Daan Fierens, Guy Van den Broeck, Ingo Thon, Bernd Gutmann, Luc De Raedt**
Department of Computer Science
Katholieke Universiteit Leuven
Celestijnenlaan 200A, 3001 Heverlee, Belgium

## Abstract

Probabilistic logic programs are logic programs in which some of the facts are annotated with probabilities. Several classical probabilistic inference tasks (such as MAP and computing marginals) have not yet received a lot of attention for this formalism. The contribution of this paper is that we develop efficient inference algorithms for these tasks. This is based on a conversion of the probabilistic logic program and the query and evidence to a weighted CNF formula. This allows us to reduce the inference tasks to well-studied tasks such as weighted model counting. To solve such tasks, we employ state-of-the-art methods. We consider multiple methods for the conversion of the programs as well as for inference on the weighted CNF. The resulting approach is evaluated experimentally and shown to improve upon the state-of-the-art in probabilistic logic programming.

## 1 Introduction

There is a lot of interest in combining probability and logic for dealing with complex relational domains. This interest has resulted in the fields of Statistical Relational Learning (SRL) and Probabilistic Logic Programming (PLP) [3]. While the two approaches essentially study the same problem, there are differences in emphasis. SRL techniques have focussed on the extension of graphical models with logical and relational representations, while PLP has extended logic programming languages (such as Prolog) with probabilities. This has resulted in differences in representation and semantics between the two approaches but also, and more importantly, in differences in the inference tasks that have been considered. The most common inference tasks in the graphical model and the SRL communities are that of computing the marginal probability of a set of random variables w.r.t. the evidence (the MARG task) and finding the most likely joint state of the random variables given the evidence (the MAP task). In the PLP community one has focussed on computing the probability of a single random variable without evidence. This paper alleviates this situation by contributing general MARG and MAP inference techniques for probabilistic logic programs.

The key contribution of this paper is a two-step approach for performing MARG and MAP inference in probabilistic logic programs. Our approach is similar to the work of Darwiche [2] and others [14, 12], who perform Bayesian network inference by conversion to weighted propositional formulae, in particular weighted CNFs. We do the same for probabilistic logic programs, a much more expressive representation framework (it extends a programming language, it allows for cycles, etc.) In the first step, the probabilistic logic program is converted to an equivalent weighted CNF. This conversion is based on well-known conversions from the knowledge representation literature. The MARG task then reduces to weighted model counting (WMC) on the resulting weighted CNF, and the MAP task to weighted MAX SAT. The second step then involves calling a state-of-the-art solver for WMC or MAX SAT. In this way, we establish new links between PLP inference and standard problems such as WMC and MAX SAT. We also identify a novel connection between PLP and Markov Logic [13].

Further contributions are made at a more technical level. First, we show how to make our approach more efficient by working only on the relevant part (with respect to query and evidence) of the given program. Second, we consider two algorithms for converting the program to a weighted CNF and compare these algorithms in terms of efficiency of the conversion process and how efficient the resulting weighted CNFs are for inference. Third, we compare the performance of different inference algorithms and show that we improve upon the state-of-the-art in PLP inference.

This paper is organized as follows. We first review the basics of LP (Section 2) and PLP (Section 3). Next we state the inference tasks that we consider (Section 4). Then we introduce our two-step approach (Section 5 and 6). Finally we evaluate this approach by means of experiments on relational data (Section 7).

## 2 Background

We now review the basics of logic programming [15] and first order logic.

### 2.1 First Order Logic (FOL)

A *term* is a variable, a constant, or a functor applied on terms. An *atom* is of the form $p(t_1, \ldots, t_n)$ where $p$ is a predicate of arity $n$ and the $t_i$ are terms. A *formula* is built out of atoms using universal and existential quantifiers and the usual logical connectives $\neg$, $\vee$, $\wedge$, $\rightarrow$ and $\leftrightarrow$. A *FOL theory* is a set of formulas that implicitly form a conjunction. An expression is called *ground* if it does not contain variables. A ground (or propositional) theory is said to be in *conjunctive normal form (CNF)* if it is a conjunction of disjunctions of literals. A *literal* is an atom or its negation. Each disjunction of literals is called a *clause*. A disjunction consisting of a single literal is called a *unit clause*. Each ground theory can be written in CNF form.

The *Herbrand base* of a FOL theory is the set of all ground atoms constructed using the predicates, functors and constants in the theory. A Herbrand interpretation, also called *(possible) world*, is an assignment of a truth value to all atoms in the Herbrand base. A world or interpretation is called a *model* of the theory if it satisfies all formulas in the theory. Satisfaction of a formula is defined in the usual way.

*Markov Logic Networks (MLNs)* [13] are a probabilistic extension of FOL. An MLN is a set of pairs of the form $(\varphi, w)$, with $\varphi$ a FOL formula and $w$ a real number called the *weight* of $\varphi$. Together with a set of constants, an MLN determines a probability distribution on the set of possible worlds. This distribution is a log-linear model (Markov random field): for every grounding of every formula $\varphi$ in the MLN, there is a feature in the log-linear model and the weight of that feature is equal to the weight of $\varphi$.

### 2.2 Logic Programming (LP)

Syntactically, a normal logic program, or briefly *logic program (LP)* is a set of rules.[1] A *rule* is a universally quantified expression of the form `h :- b1, ... , bn`, where $h$ is an atom and the $b_i$ are literals. The atom $h$ is called the *head* of the rule and $b_1, \ldots, b_n$ the *body*, representing the conjunction $b_1 \wedge \ldots \wedge b_n$. A *fact* is a rule that has *true* as its body and is written more compactly as `h`. Note that ':-' can also be written as '$\leftarrow$'. Hence, each rule can syntactically be seen as a FOL formula. There is a crucial difference in semantics, however.

We use the *well-founded semantics* for LPs. In the case of a negation-free LP (a 'definite' program), the well-founded model is identical to the *Least Herbrand Model (LHM)*. The LHM is defined as the least ('smallest') of all models obtained when interpreting the LP as a FOL theory. The *least* model is the model that is a subset of all other models (in the sense that it makes the fewest atoms true). Intuitively, the LHM is the set of all ground atoms that are entailed by the LP. For negation-free LPs, the LHM is guaranteed to exist and be unique. For LPs with negation, we use the well-founded model, see [15].

The reason why one considers only the *least* model of an LP is that LP semantics makes the *closed world assumption (CWA)*. Under the CWA, everything that is not certainly true is assumed to be false. This has implications on how to interpret rules. Given a ground LP and an atom $a$, the set of all rules with $a$ in the head should be read as the *definition* of $a$: the atom $a$ is defined to be true if and only if at least one of the rule bodies is true (the 'only if' is due to the CWA).

### 2.3 Differences between FOL and LP

There is a crucial difference in semantics between LP and FOL: LP makes the CWA while FOL does not. For example, the FOL theory $\{a \leftarrow b\}$ has 3 models $\{\neg a, \neg b\}$, $\{a, \neg b\}$ and $\{a, b\}$. The syntactically equivalent LP `{a :- b}` has only one model, namely the least Herbrand model $\{\neg a, \neg b\}$ (intuitively, $a$ and $b$ are false because there is no rule that makes $b$ true, and hence there is no applicable rule that makes $a$ true either).

Because LP is syntactically a subset of FOL, it is tempting to believe that FOL is more 'expressive' than LP. This is wrong because of the difference in semantics. In the knowledge representation literature, it has been shown that certain concepts that can be expressed in (non-ground) LP cannot be expressed in (non-ground) FOL, for instance inductive definitions [5]. This motivates our interest in LP and PLP.

## 3 Probabilistic Logic Programming

Most probabilistic programming languages, including PRISM [3], ICL [3], ProbLog [4] and LPAD [11], are based on Sato's *distribution semantics* [3]. In this paper we use ProbLog as it is the simplest of these languages. However, our approach can easily be used for the other languages as well.

---
[1] Rules are also called *normal clauses* but we avoid this terminology because we use 'clause' in the context of CNF.

**Syntax.** A ProbLog program consists of a set of probabilistic facts and a logic program, i.e. a set of rules. A *probabilistic fact*, written `p::f`, is a fact $f$ annotated with a probability $p$. An atom that unifies with a probabilistic fact is called a *probabilistic atom*, while an atom that unifies with the head of some rule is called a *derived atom*. The set of probabilistic atoms must be disjoint from the set of derived atoms. Below we use as an example a ProbLog program with probabilistic facts `0.3::rain` and `0.2::sprinkler` and rules `wet :- rain` and `wet :- sprinkler`. Intuitively, this program states that it rains with probability 0.3, the sprinkler is on with probability 0.2, and the grass is wet if and only if it rains or the sprinkler is on. Compared to PLP languages like PRISM and ICL, ProbLog is less restricted with respect to the rules that are allowed in a program. PRISM and ICL require the rules to be acyclic (or contingently acyclic) [3]. In addition, PRISM requires rules with unifiable heads to have mutually exclusive bodies (such that at most one of these bodies is true at once; this is the mutual exclusiveness assumption). ProbLog does not have these restrictions, for instance, we can have cyclic programs with rules such as `smokes(X) :- friends(X,Y), smokes(Y)`. This type of cyclic rules are often needed for tasks such as collective classification or social network analysis.

**Semantics.** A ProbLog program specifies a probability distribution over possible worlds. To define this distribution, it is easiest to consider the grounding of the program with respect to the Herbrand base. Each ground probabilistic fact `p::f` gives an *atomic choice*, i.e. we can choose to include $f$ as a fact (with probability $p$) or discard it (with probability $1 - p$). A *total choice* is obtained by making an atomic choice for each ground probabilistic fact. To be precise, a total choice is any subset of the set of all ground probabilistic atoms. Hence, if there are $n$ ground probabilistic atoms then there are $2^n$ total choices. Moreover, we have a probability distribution over these total choices: the probability of a total choice is defined to be the product of the probabilities of the atomic choices that it is composed of (atomic choices are seen as independent events). In our above example, there are 4 total choices: $\{\}, \{rain\}, \{sprinkler\}$, and $\{rain, sprinkler\}$. The probability of the total choice $\{rain\}$, for instance, is $0.3 \times (1 - 0.2)$.

Given a particular total choice $C$, we obtain a logic program $C \cup R$, where $R$ denotes the rules in the ProbLog program. This logic program has exactly one well-founded model[2] $WFM(C \cup R)$. We call a given world $\omega$ a *model* of the ProbLog program if there indeed exists a total choice $C$ such that $WFM(C \cup R) = \omega$. We use $MOD(L)$ to denote the set of all models of a ProbLog program $L$. In our example, the total choice $\{rain\}$ yields the logic program $\{$`rain, wet :- rain, wet :- sprinkler`$\}$. The WFM of this program is the world $\{rain, \neg sprinkler, wet\}$. Hence this world is a model. There are three more models corresponding to each of the three other total choices. An example of a world that is not a model of the ProbLog program is $\{rain, \neg sprinkler, \neg wet\}$ (it is impossible that *wet* is false while *rain* is true).

Everything is now in place to define the distribution over possible worlds: the probability of a world that is a model of the ProbLog program is equal to the probability of its total choice; the probability of a world that is not a model is 0. For example, the probability of the world $\{rain, \neg sprinkler, wet\}$ is $0.3 \times (1 - 0.2)$, while the probability of $\{rain, \neg sprinkler, \neg wet\}$ is 0.

## 4 Inference Tasks

Let **At** be the set of all ground (probabilistic and derived) atoms in a given ProbLog program. We assume that we are given a set $\mathbf{E} \subset \mathbf{At}$ of observed atoms (*evidence atoms*), and a vector **e** with their observed truth values (i.e. the evidence is $\mathbf{E} = \mathbf{e}$). We are also given a set $\mathbf{Q} \subset \mathbf{At}$ of atoms of interest (*query atoms*). The two inference tasks that we consider are MARG and MAP. MARG is the task of computing the marginal distribution of every query atom given the evidence, i.e. computing $P(Q \mid \mathbf{E} = \mathbf{e})$ for each $Q \in \mathbf{Q}$. MAP (maximum a posteriori) is the task of finding the most likely joint state of all query atoms given the evidence, i.e. finding $argmax_{\mathbf{q}} P(\mathbf{Q} = \mathbf{q} \mid \mathbf{E} = \mathbf{e})$.

**Existing work.** In the literature on probabilistic graphical models and statistical relational learning, MARG and MAP have received a lot of attention, while in PLP the focus has been on the special case of MARG where there is a single query atom ($\mathbf{Q} = \{Q\}$) and no evidence ($\mathbf{E} = \emptyset$). This task is often referred to as computing the *success probability* of $Q$ [4]. The only works related to the more general MARG or MAP task in the PLP literature [3, 11, 8] make a number of restrictive assumptions about the given program such as acyclicity [8] and the mutual exclusiveness assumption (for PRISM [3]). There also exist approaches that transform ground probabilistic programs to Bayesian networks and then use standard Bayesian network inference procedures [11]. However, these are also re-stricted to being acyclic and in addition they work only for already grounded programs. Our approach does not suffer from such restrictions and is applicable to all ProbLog programs. It consists of two steps: 1) conversion of the program to a weighted CNF and 2) inference on the resulting weighted CNF. We discuss these two steps in the next sections.

---

[2]Some LPs have a three-valued WFM (atoms are true, false or unknown), but we consider only ProbLog programs for which all LPs are two-valued (no unknowns) [15].

## 5 Conversion to Weighted CNF

The following algorithm outlines how we convert a ProbLog program $L$ together with a query $\mathbf{Q}$ and evidence $\mathbf{E} = \mathbf{e}$ to a weighted CNF:

1. *Ground $L$ yielding a program $L_g$ while taking into account $\mathbf{Q}$ and $\mathbf{E} = \mathbf{e}$ (cf. Theorem 1, Section 5.1).*
   It is unnecessary to consider the full grounding of the program, we only need the part that is relevant to the query given the evidence, that is, the part that captures the distribution $P(\mathbf{Q} \mid \mathbf{E} = \mathbf{e})$. We refer to the resulting program $L_g$ as the *relevant ground program* with respect to $\mathbf{Q}$ and $\mathbf{E} = \mathbf{e}$.

2. *Convert the ground rules in $L_g$ to an equivalent CNF $\varphi_r$ (cf. Lemma 1, Section 5.2).*
   This step takes into account the logic programming semantics of the rules in order to generate an equivalent CNF formula.

3. *Define a weight function for all atoms in $\varphi = \varphi_r \wedge \varphi_e$ (cf. Theorem 2, Section 5.3).*
   To obtain the weighted CNF, we first condition on the evidence by defining the CNF $\varphi$ as the conjunction of the CNF $\varphi_r$ for the rules and the evidence $\varphi_e$. Then we define the weight function for $\varphi$.

The correctness of the algorithm is shown below; this relies on the indicated theorems and lemma's. Before describing each of the steps in detail in the following subsections, we illustrate the algorithm on our simple example ProbLog program, with probabilistic facts `0.3::r` and `0.2::s` and rules `w :- r` and `w :- s` (we abbreviate *rain* to $r$, etc.). Suppose that the query set $\mathbf{Q}$ is $\{r, s\}$ and the evidence is that $w$ is false. Step 1 finds the relevant ground program. Since the program is already ground and all parts are relevant here, this is simply the program itself. Step 2 converts the rules in the relevant ground program to an equivalent CNF. The resulting CNF $\varphi_r$ contains the following three clauses (see Section 5.2): $\neg r \vee w$, $\neg s \vee w$, and $\neg w \vee s \vee r$. Step 3 conditions $\varphi_r$ on the evidence. Since we have only one evidence atom in our example ($w$ is false), all we need to do is to add the unit clause $\neg w$ to the CNF $\varphi_r$. The resulting CNF $\varphi$ is $\varphi_r \wedge \neg w$. Step 3 also defines the weight function, which assigns a weight ($\in [0, 1]$) to each literal in $\varphi$; see Section 5.3. This results in the *weighted CNF*, that is, the combination of the weight function and the CNF $\varphi$.

### 5.1 The Relevant Ground Program

In order to convert the ProbLog program to CNF we first need to ground it.[3] We try to find the part of the grounding that is relevant to the queries $\mathbf{Q}$ and the evidence $\mathbf{E} = \mathbf{e}$. To do so, we make use of the concept of a dependency set with respect to a ProbLog program [8].

The *dependency set* of a ground atom $a$ is the set of ground atoms that occur in a proof of $a$. The dependency set of multiple atoms is the union of their dependency sets. We call a ground atom *relevant* with respect to $\mathbf{Q}$ and $\mathbf{E}$ if it occurs in the dependency set of $\mathbf{Q} \cup \mathbf{E}$. Similarly, we call a ground rule relevant if it contains relevant ground atoms. It is safe to restrict the grounding to the relevant rules only [8]. To find these rules we apply SLD resolution to prove all atoms in $\mathbf{Q} \cup \mathbf{E}$ (this can be seen as backchaining over the rules starting from $\mathbf{Q} \cup \mathbf{E}$). We employ memoization to avoid proving the same atom twice (and to avoid going into an infinite loop if the rules are cyclic). The relevant rules are simply all ground rules encountered during the resolution process.

The above grounding algorithm does not make use of all the information about the evidence $\mathbf{E} = \mathbf{e}$. Concretely, it makes use of which atoms are in the evidence ($\mathbf{E}$) but not of what their value is ($\mathbf{e}$). We can make use of this as well. Call a ground rule *inactive* if the body of the rule contains a literal $l$ that is false in the evidence ($l$ can be an atom that is false in $\mathbf{e}$, or the negation of an atom that is true in $\mathbf{e}$). Inactive rules do not contribute to the semantics of a program. Hence they can be omitted. In practice, we do this simultaneously with the above process: we omit inactive rules encountered during the SLD resolution.

The result of this grounding algorithm is what we call the *relevant ground program* $L_g$ for $L$ with respect to $\mathbf{Q}$ and $\mathbf{E} = \mathbf{e}$. It contains all the information necessary for solving MARG or MAP about $\mathbf{Q}$ given $\mathbf{E} = \mathbf{e}$. The advantage of this 'focussed' approach is that the weighted CNF becomes more compact, which makes subsequent inference more efficient. The disadvantage is that we need to redo the conversion to weighted CNF when the evidence and queries change. This is no problem since the conversion is fast compared to the actual inference (see Section 7).

---

[3] In Section 2.3 we stated that some non-ground LPs cannot be expressed in non-ground FOL. In contrast, each ground LP can be converted to an equivalent ground FOL formula or CNF [9].

**Theorem 1** *Let $L$ be a ProbLog program and let $L_g$ be the relevant ground program for $L$ with respect to $\mathbf{Q}$ and $\mathbf{E} = \mathbf{e}$. $L$ and $L_g$ specify the same distribution $P(\mathbf{Q} \mid \mathbf{E} = \mathbf{e})$.*

The proofs of all theorems can be found in a technical report [6].

### 5.2 The CNF for the Ground Program

We now discuss how to convert the rules in $L_g$ to an equivalent CNF $\varphi_r$. For this conversion, the following lemma holds; it will be used in the next section.

**Lemma 1** *Let $L_g$ be a ground ProbLog program. Let $\varphi_r$ denote the CNF derived from the rules in $L_g$. Then $SAT(\varphi_r) = MOD(L_g)$.*[4]

Recall that $MOD(L_g)$ denotes the set of models of a ProbLog program $L_g$ (Section 3). On the CNF side, $SAT(\varphi_r)$ denote the set of models of a CNF $\varphi_r$.

Converting a set of logic programming (LP) rules to an equivalent CNF is a purely logical (non-probabilistic) problem and has been well studied in the LP literature (e.g. [9]). Since the problem is of a highly technical nature, we are unable to repeat the full details in the present paper, but shall refer to the literature for more details. Note that converting LP rules to CNF is not simply a syntactical matter of rewriting the rules in the appropriate form. The point is that the rules and the CNF are to be interpreted according to a different semantics (LP versus FOL, recall Section 2). Hence the conversion should compensate for this: the rules under LP semantics (with Closed World Assumption) should be equivalent to the CNF under FOL semantics (without CWA).

For *acyclic rules*, the conversion is straightforward, we simply take *Clark's completion* of the rules [9, 8]. For instance, consider the rules `w :- r` and `w :- s`. Clark's completion of these rules is the FOL formula $w \leftrightarrow r \lor s$. Once we have the FOL formula, obtaining the CNF is simply a rewriting issue. For our example, we obtain a CNF with three clauses: $\neg r \lor w$, $\neg s \lor w$, and $\neg w \lor s \lor r$ (the last clause reflects the CWA).[5]

For *cyclic rules*, the conversion is more complicated. This holds in particular for rules with 'positive' loops (atoms depend positively on each other, e.g. `smokes(X) :- friends(X,Y), smokes(Y)`). It is well known that for such rules Clark's completion is not correct, i.e. the resulting CNF is not equivalent to the rules [9]. A range of more sophisticated conversion algorithms have been developed. We use two such algorithms. Given a set of rules, both algorithms derive an equivalent CNF (that satisfies Lemma 1). The CNFs generated by the two algorithms might be syntactically different because the algorithms introduce a set of auxiliary atoms in the CNF and these sets might differ. For both algorithms, the size of the CNF typically increases with the 'loopyness' of the given rules. We now briefly discuss both algorithms.

**Rule-based conversion to CNF.** The first algorithm [9] belongs to the field of Answer Set Programming. It first rewrites the given rules into an equivalent set of rules without positive loops (all resulting loops involve negation). This requires the introduction of auxiliary atoms and rules. Since the resulting rules are free of positive loops, they can be converted by simply taking Clark's completion. The result can then be written as a CNF.

**Proof-based conversion to CNF.** The second algorithm [10] is proof-based. It first constructs all proofs of all atoms of interest, in our case all atoms in $\mathbf{Q} \cup \mathbf{E}$, using tabled SLD resolution. The proofs are collected in a recursive structure (a set of 'nested tries' [10]), which will have loops if the given rules had loops. The algorithm then operates on this structure in order to 'break' the loops and obtain an equivalent Boolean formula. This formula can then be written as a CNF.

### 5.3 The Weighted CNF

The final step constructs the weighted CNF from the CNF $\varphi_r$. First, the CNF $\varphi$ is defined as the conjunction of $\varphi_r$ and a CNF $\varphi_e$ capturing the evidence $\mathbf{E} = \mathbf{e}$. Here $\varphi_e$ is a conjunction of unit clauses (there is a unit clause $a$ for each true atom and a clause $\neg a$ for each false atom in the evidence). Second, we define the weight function for all literals in the resulting CNF. If the ProbLog program contains a probabilistic fact `p::f`, then we assign weight $p$ to $f$ and weight $1-p$ to $\neg f$. Derived literals (literals not occuring in a probabilistic fact) get weight 1. In our example, we had two probabilistic facts `0.3::r` and `0.2::s`, and one derived atom $w$. The weight function is $\{r \mapsto 0.3, \neg r \mapsto 0.7, s \mapsto 0.2, \neg s \mapsto 0.8, w \mapsto 1, \neg w \mapsto 1\}$. The *weight of a world* $\omega$, denoted $w(\omega)$, is defined to be the product of the weight of all literals in $\omega$. For example, the world $\{r, \neg s, w\}$ has weight $0.3 \times 0.8 \times 1$.

We have now seen how to construct the entire weighted CNF from the relevant ground program. The following theorem states that this weighted CNF is equivalent - in some sense - to the relevant ground program. We will make use of this result when performing inference on the weighted CNF.

---

[4] The conversion from rules to CNF $\varphi_r$ sometimes introduces additional atoms. We can safely omit these atoms from the models in $SAT(\varphi_r)$ because their truth value is uniquely defined by the truth values of the original atoms (w.r.t. the original atoms: $SAT(\varphi_r) = MOD(L_g)$).

[5] For ProbLog programs encoding Boolean Bayesian networks the resulting CNF equals that of Sang et al. [14].

**Theorem 2** *Let $L_g$ be the relevant ground program for some ProbLog program with respect to $\mathbf{Q}$ and $\mathbf{E} = \mathbf{e}$. Let $MOD_{\mathbf{E}=\mathbf{e}}(L_g)$ be those models in $MOD(L_g)$ that are consistent with the evidence $\mathbf{E} = \mathbf{e}$. Let $\varphi$ denote the CNF and $w(.)$ the weight function of the weighted CNF derived from $L_g$. Then:*
*(model equivalence) $SAT(\varphi) = MOD_{\mathbf{E}=\mathbf{e}}(L_g)$,*
*(weight equivalence) $\forall \omega \in SAT(\varphi): w(\omega) = P_{L_g}(\omega)$, i.e. the weight of $\omega$ according to $w(.)$ is equal to the probability of $\omega$ according to $L_g$.*

Note the relationship with Lemma 1: while Lemma 1 applies to the CNF $\varphi_r$ *prior* to conditioning on the evidence, Theorem 2 applies to the CNF $\varphi$ *after* conditioning.

The weighted CNF can also be regarded as a ground *Markov Logic Network (MLN)*. The MLN contains all clauses that are in the CNF (as 'hard' clauses) and also contains two weighted unit clauses per probabilistic atom. For example, for a probabilistic atom $r$ and weight function $\{r \mapsto 0.3, \neg r \mapsto 0.7\}$, the MLN contains a unit clause $r$ with weight $ln(0.3)$ and a unit clause $\neg r$ with weight $ln(0.7)$.[6] We have the following equivalence result.

**Theorem 3** *Let $L_g$ be the relevant ground program for some ProbLog program with respect to $\mathbf{Q}$ and $\mathbf{E} = \mathbf{e}$. Let $\mathcal{M}$ be the corresponding ground MLN. The distribution $P(\mathbf{Q})$ according to $\mathcal{M}$ is the same as the distribution $P(\mathbf{Q} \mid \mathbf{E} = \mathbf{e})$ according to $L_g$.*

Note that for the MLN we consider the distribution $P(\mathbf{Q})$ (not conditioned on the evidence). This is because the evidence is already hard-coded in the MLN.

## 6 Inference on Weighted CNFs

To solve the MARG and MAP inference tasks for the original probabilistic logic program $L$, the query $\mathbf{Q}$ and evidence $\mathbf{E} = \mathbf{e}$, we have converted the program to a weighted CNF. A key advantage is that the original MARG and MAP inference tasks can now be reformulated in terms of well-known tasks such as weighted model counting on the weighted CNF. This implies that we can use any of the existing state-of-the-art algorithms for solving these tasks. In other words, by the conversion of ProbLog to weighted CNF, we "get the inference algorithms for free".

### 6.1 MARG Inference

Let us first discuss how to tackle MARG, the task of computing the marginal $P(Q \mid \mathbf{E} = \mathbf{e})$ for each query atom $Q \in \mathbf{Q}$, and its special case 'MARG-1' where $\mathbf{Q}$ consists of a single query atom ($\mathbf{Q} = \{Q\}$).

**1) Exact/approximate MARG-1 by means of weighted model counting.** By definition, $P(Q = q \mid \mathbf{E} = \mathbf{e}) = P(Q = q, \mathbf{E} = \mathbf{e})/P(\mathbf{E} = \mathbf{e})$. The denominator is equal to the *weighted model count* of the weighted CNF $\varphi$, namely $\sum_{\omega \in SAT(\varphi)} w(\omega)$.[7] Similarly, the numerator is the weighted model count of the weighted CNF $\varphi_q$ obtained by conjoining the original weighted CNF $\varphi$ with the appropriate unit clause for $Q$ (namely $Q$ if $q = true$ and $\neg Q$ if $q = false$). Hence each marginal can be computed by solving two weighted model counting (WMC) instances. WMC is a well-studied task in the SAT community. Solving these WMC instances can be done using any of the existing algorithms (exact [1] or approximate [7]).

It is well-known that MARG inference with Bayesian networks can be solved using WMC [14]. This paper is the first to point out that this also holds for inference with probabilistic logic programs. The experiments below show that this approach improves upon state-of-the-art methods in probabilistic logic programming.

**2) Exact MARG by means of compilation.** To solve the general MARG task with multiple query atoms, one could simply solve each of the MARG-1 tasks separately using WMC as above. However, this would lead to many redundant computations. A popular solution to avoid this is to first *compile* the weighted CNF into a more efficient representation [2]. Concretely, we can compile the CNF to **d-DNNF** (deterministic Decomposable Negation Normal Form [1]) and then compute all required marginals from the (weighted) d-DNNF. The latter can be done efficiently for all marginals in parallel, namely by traversing the d-DNNF twice [2].[8]

In the probabilistic logic programming (PLP) commu-nity, the state-of-the-art is to compile the program into another form, namely a **BDD** (reduced ordered Binary Decision Diagram) [4]. The BDD approach has recently also been used for MARG inference (to compute all marginals, the BDD is then traversed in a way that is very similar to that for d-DNNFs [8]). BDDs form a subclass of d-DNNFs [1]. So far, general d-DNNFs have not been considered in the PLP community, despite the theoretical and empirical evidence that compilation to d-DNNF outperforms compilation to BDD in the context of model counting [1]. Our experimental results (Section 7) confirm the superiority of d-DNNFs.

---

[6] A 'hard' clause has weight infinity (each world that violates the clause has probability zero). The logarithms, e.g. $ln(0.3)$, are needed because an MLN is a log-linear model. The logarithms are negative, but any MLN with negative weights can be rewritten into an equivalent MLN with only positive weights [3].

[7] This is because $P(\mathbf{E} = \mathbf{e}) = \sum_{\omega \in MOD_{\mathbf{E}=\mathbf{e}}(L)} P_L(\omega) = \sum_{\omega \in SAT(\varphi)} w(\omega)$, where the second equality follows from Theorem 2 (model equivalence implies that the sets of over which the sums range are equal; weight equivalence implies that the summed terms are equal).

[8] In the literature one typically converts the weighted d-DNNF to an arithmetic circuit (AC) and then traverses this AC. This is equivalent to our approach (the conversion to AC is not strictly necessary, we sidestep it).

**3) Approximate MARG by means of MCMC.** We can also use sampling (MCMC) on the weighted CNF. Because the CNF itself is deterministic, standard MCMC approaches like Gibbs sampling are not suited. We use the *MC-SAT* algorithm that was developed specifically to deal with determinism (in each step of the Markov chain, MC-SAT makes use of a SAT solver to construct a new sample) [13]. MC-SAT was developed for MLNs. Theorem 3 ensures that MCMC on the appropriate MLN samples from the correct distribution $P(\mathbf{Q} \mid \mathbf{E} = \mathbf{e})$.

### 6.2 MAP Inference

Also MAP inference on weighted CNFs has been studied before. We consider the following algorithms.

**1) Exact MAP by means of compilation.** We can compile the weighted CNF to a weighted d-DNNF and then use this d-DNNF to find the MAP solution, see Darwiche [2]. The compilation phase is in fact independent of the specific task (MARG or MAP), only the traversal differs. Compilation to BDD is also possible.

**2) Approximate MAP/MPE by means of stochastic local search.** MPE is the special case of MAP where one wants to find the state of *all* nonevidence atoms. MPE inference on a weighted CNF reduces to the weighted MAX SAT problem [12], a standard problem in the SAT literature. A popular approximate approach is stochastic local search [12]. An example algorithm is *MaxWalkSAT*, which is also the standard MPE algorithm for MLNs [3].

## 7 Experiments

Our implementation currently supports **(1)** exact MARG by compilation to either d-DNNF or BDD, **(2)** approximate MARG with MC-SAT, **(3)** approximate MAP/MPE with MaxWalkSAT. Other algorithms for inference on the weighted CNF could be applied as well, so the above list is not exhaustive.

The goal of our experiments is to establish the feasibility of our approach and to analyze the influence of the different parameters. We focus on MARG inference (for MAP/MPE, our current implementation is a proof-of-concept).

### 7.1 Experimental Setup

**Domains.** As a *social network* domain we use the standard 'Smokers' domain [3]. The main rule in the ProbLog program is `smokes(X) :- friend(X,Y), smokes(Y), inf(Y,X)`. There is also a probabilistic fact `p::inf(X,Y)`, which states that $Y$ influences $X$ with probability $p$ (for each ground $(X, Y)$ pair there is an independent atomic choice). This means that each smoking friend $Y$ of $X$ independently *causes* $X$ to smoke with probability $p$. Other rules state that people smoke for other reasons as well, that smoking causes cancer, etc. All probabilities in the program were set manually.

We also use the WebKB dataset, a *collective classification* domain (http://www.cs.cmu.edu/~webkb/). In WebKB, university webpages are to be tagged with classes (e.g. 'course page'). This is modelled with a predicate $hasclass(P, Class)$. The rules specify how the class of a page $P$ depends on the textual content of $P$, and on the classes of pages that link to $P$. All probabilities were learned from data [8].

**Inference tasks.** We vary the domain size (number of people/pages). For each size we use 8 different instances of the MARG task and report median results. We generate each instance in 3 steps. **(1)** We generate the network. For WebKB we select a random set of pages and use the link structure given in the data. For Smokers we use power law random graphs since they are known to resemble social networks. **(2)** We select query and evidence atoms (**Q** and **E**). For WebKB we use half of all *hasclass* atoms as query and the other half as evidence. For Smokers we do the same with all $friends$ and $cancer$ atoms. **(3)** We generate a sample of the ProbLog program to determine the value **e** of the evidence atoms. For every query atom we also sample a truth value; we store these values and use them later as 'query ground truth' (see Section 7.3).

### 7.2 Influence of the Grounding Algorithm

We compare computing the relevant ground program (RGP) with naively doing the complete grounding.

**Grounding.** The idea behind the RGP is to reduce the grounding by pruning clauses that are irrelevant or inactive w.r.t. the queries and evidence. Our setup is such that all clauses are relevant. Hence, the only reduction comes from pruning inactive clauses (that have a false evidence literal in the body). The effect of this pruning is small: on average the size of the ground program is reduced by 17% (results not shown).

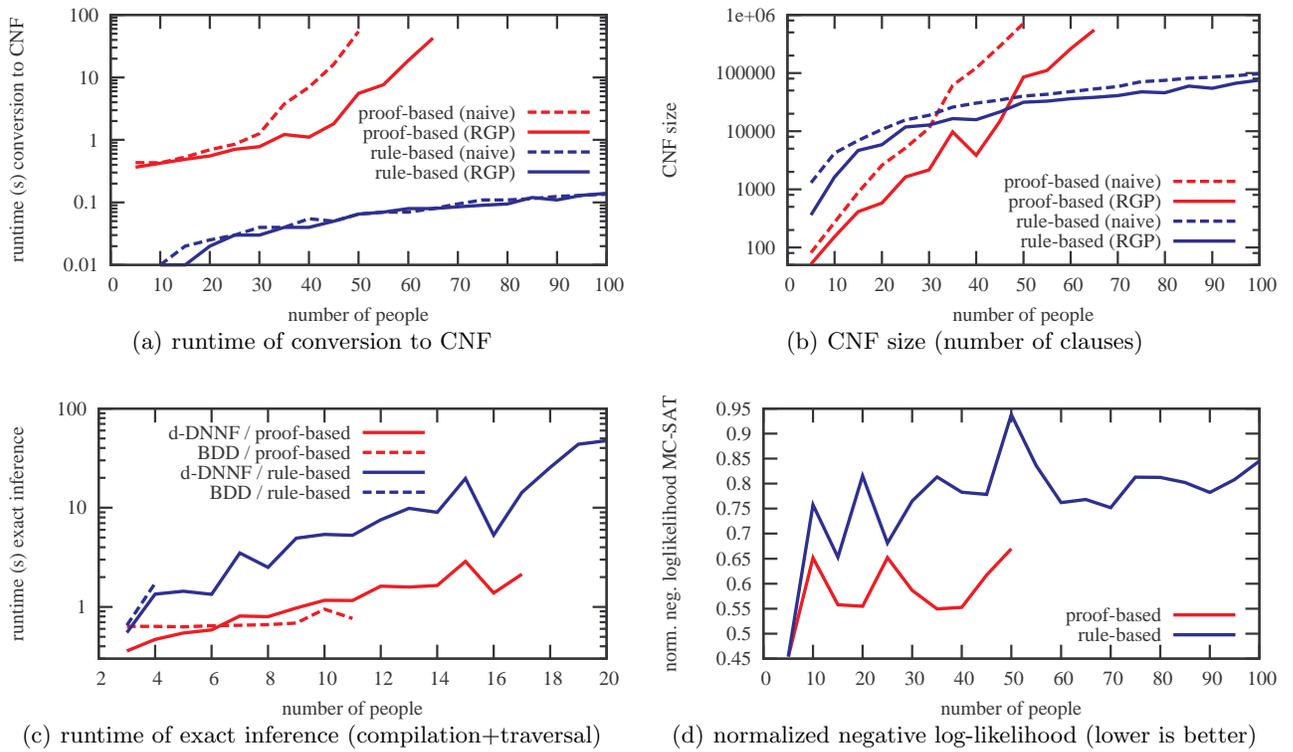

Figure 1: Results for *Smokers* in function of domain size. (When the curve for an algorithm ends at a particular domain size, this means that the algorithm is intractable beyond that size.)

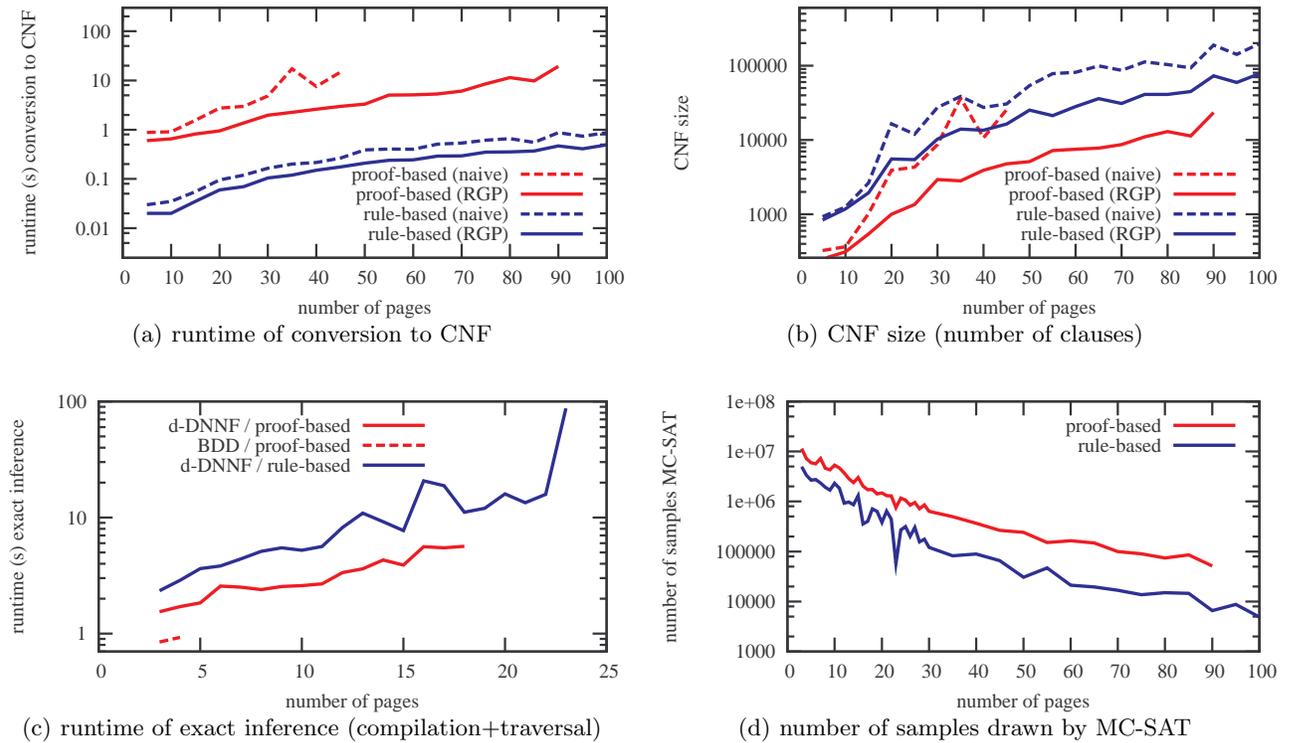

Figure 2: Results for *WebKB* in function of domain size.

**Implications on the conversion to CNF.** The proof-based conversion becomes intractable for large domain sizes, but the size where this happens is significantly larger when working on the RGP instead of on the complete grounding (see Fig. 1a/2a). Also the size of the CNFs is reduced significantly by using the RGP (up to a 90% reduction, Fig. 1b/2b). The reason why a 17% reduction of the program can yield a 90% reduction of the CNF is that loops in the program cause a 'blow-up' of the CNF. Removing only a few rules in the ground program can already break loops and make the CNF significantly smaller. Note that the proof-based conversion suffers from this blow-up more than the rule-based conversion does.

Computing the grounding is always very fast, both for the RGP and the complete grounding (milliseconds on Smokers; around 1s for WebKB). We conclude that using the RGP instead of the complete grounding is beneficial and comes at almost no computational cost. Hence, from now on we always use the RGP.

### 7.3 Influence of the Conversion Algorithm

We compare the rule-based and proof-based algorithm for converting ground rules to CNF (Section 5.2).

**Conversion.** The proof-based algorithm, by its nature, does more effort to convert the program into a compact CNF. This has implication on the scalability of the algorithm: on small domains the algorithm is fast, but on larger domains it becomes intractable (Fig. 1a/2a). In contrast, the rule-based algorithm is able to deal with all considered domain sizes and is always fast (runtime at most 0.5s). A similar trend holds in terms of CNF size. For small domains, the proof-based algorithm generates the smallest CNFs, but for larger domains the opposite holds (Fig. 1b/2b).

**Implications on inference.** We discuss the influence of the conversion algorithm on exact inference in the next section. Here we focus on approximate inference. We use MC-SAT as a tool to evaluate how efficient the different CNFs are for inference. Concretely, we run MC-SAT on the two types of CNFs and measure the quality of the estimated marginals. Evaluating the quality of approximate marginals is non-trivial when computing true marginals is intractable. We use the same solution as the original MC-SAT paper: we let MC-SAT run for a fixed time (10 minutes) and measure the quality of the estimated marginals as the likelihood of the 'query ground truth' according to these estimates (see [13] for the motivation).

On domain sizes where the proof-based algorithm is still tractable, inference results are better with the proof-based CNFs than with the rule-based CNFs (Fig. 1d). This is because the proof-based CNFs are more compact and hence more samples can be drawn in the given time (Fig. 2d).

We conclude that for smaller domains the proof-based algorithm is preferable because of the smaller CNFs. On larger domains, the rule-based algorithm should be used.

### 7.4 Influence of the Inference Algorithm

We focus on the comparison of the two exact inference algorithms, namely compilation to d-DNNFs or BDDs. We make the distinction between inference on rule-based and proof-based CNFs (in the PLP literature, BDDs have almost exclusively been used for proof-based CNFs [4, 8]).[9]

**Proof-based CNFs.** On the Smokers domain, BDDs perform relatively well, but they are nevertheless clearly outperformed by the d-DNNFs (Fig. 1c). On WebKB, the difference is even larger: BDDs are only tractable on domains of size 3 or 4, while d-DNNFs reach up to size 18 (Fig. 2c). When BDDs become intractable, this is mostly due to memory problems.[10]

**Rule-based CNFs.** These CNFs are less compact than the proof-based CNFs (at least for those domain sizes where exact inference is feasible). The results clearly show that the d-DNNFs are much better at dealing with these CNFs than the BDDs are. Concretely, the d-DNNFs are still tractable up to reasonable sizes. In contrast, using BDDs on these rule-based CNFs is nearly impossible: on Smokers the BDDs only solve size 3 and 4, on WebKB they even do not solve any of the inference tasks on rule-based CNFs.

We conclude that the use of d-DNNFs pushes the limit of exact MARG inference significantly further as compared to BDDs, which are the state-of-the-art in PLP.

## 8 Conclusion

This paper contributes a two-step procedure for MAP and MARG inference in general probabilistic logic programs. The first step generates a weighted CNF that captures all relevant information about a specific query, evidence and probabilistic logic program. This step relies on well-known conversion techniques from logic programming. The second step then invokes well-known solvers (for instance for WMC and weighted MAX SAT) on the generated weighted CNF.

---

[9]Compiling our proof-based CNFs to BDDs yields exactly the same BDDs as used by Gutmann et al. [8]. In the special case of a single query and no evidence, this also equals the BDDs used by De Raedt et al. [4].

[10]It might be surprising that BDDs, which are the state-of-the-art in PLP, do not perform better. However, one should keep in mind that we are using BDDs for *exact* inference here. BDDs are also used for approximate inference, one simply compiles an *approximate CNF* into a BDD [4]. The same can be done with d-DNNFs, and we again expect improvement over BDDs.

Our two-step approach is akin to that employed in the Bayesian network community where many inference problems are also cast in terms of weighted CNFs [2, 12, 14]. We do the same for probabilistic logic programs, which are much more expressive (as they extend a programming language and do not need to be acyclic). This conversion-based approach is advantageous because it allows us to employ a wide range of well-known and optimized solvers on the weighted CNFs, essentially giving us "inference algorithms for free". Furthermore, the approach also improves upon the state-of-the-art in probabilistic logic programming, where one has typically focussed on inference with a single query atom and no evidence (cf. Section 4), often by using BDDs. By using d-DNNFs instead of BDDs, we obtained speed-ups that push the limit of exact MARG inference significantly further.

Our approach also provides new insights into the relationships between PLP and other frameworks. As one immediate outcome, we pointed out a conversion of probabilistic logic programs to ground Markov Logic, which allowed us to apply MC-SAT to PLP inference. This contributes to further bridging the gap between PLP and the field of statistical relational learning.

### Acknowledgements

DF, GVdB and BG are supported by the Research Foundation-Flanders (FWO-Vlaanderen). Research supported by the European Commission under contract number FP7-248258-First-MM. We thank Maurice Bruynooghe, Theofrastos Mantadelis and Kristian Kersting for useful discussions.